\documentclass[preprint,12pt]{elsarticle}
\biboptions{numbers,sort&compress}
\usepackage[T1]{fontenc}
\usepackage{amsmath,amssymb,amsthm,mathtools}
\usepackage{graphicx,booktabs,array,tabularx,multirow}
\usepackage{microtype}
\usepackage[hidelinks]{hyperref}
\graphicspath{{./}}
\newtheorem{theorem}{Theorem}
\newtheorem{proposition}{Proposition}
\newtheorem{lemma}{Lemma}
\newtheorem{corollary}{Corollary}

\newcommand{\T}{\mathcal{T}}
\newcommand{\W}{\mathcal{W}}
\newcommand{\R}{\mathbb{R}}
\newcommand{\B}{\mathbb{B}}
\newcommand{\Phys}{\operatorname{Phys}}

\newcommand{\conv}{\operatorname{conv}}

\newcommand{\norm}[1]{\left\lVert #1\right\rVert}

\newcolumntype{Y}{>{\raggedright\arraybackslash}X}
\hypersetup{pdftitle={Planning to Stop: Factored Certification and Terminal Repair for Noncircular Robots},pdfauthor={Runtian Zheng; Congpeng Zhang; Ying Liu}}
\newcommand{\primaryrows}{%
Portal-only & 27/139 (19.4) & 83/139 (59.7) & 29/118 (24.6) & 39/118 (33.1) \\
Stop-set & 44/139 (31.7) & 97/139 (69.8) & 34/118 (28.8) & 49/118 (41.5) \\
Stop-set + repair & 50/139 (36.0) & 98/139 (70.5) & 31/118 (26.3) & 50/118 (42.4) \\
Factored, no repair & 55/139 (39.6) & 102/139 (73.4) & 37/118 (31.4) & 61/118 (51.7) \\
GeoDiff & 61/139 (43.9) & 102/139 (73.4) & 37/118 (31.4) & 59/118 (50.0) \\
Smac Hybrid B & 1/139 (0.7) & 6/139 (4.3) & 0/118 (0.0) & 2/118 (1.7) \\
Smac Hybrid B+ & 2/139 (1.4) & 6/139 (4.3) & 0/118 (0.0) & 2/118 (1.7) \\
Smac Lattice B & 9/139 (6.5) & 25/139 (18.0) & 3/118 (2.5) & 9/118 (7.6) \\
Smac Lattice B+ & 13/139 (9.4) & 26/139 (18.7) & 3/118 (2.5) & 10/118 (8.5) \\
DDR-opt B & 6/139 (4.3) & 10/139 (7.2) & 1/118 (0.8) & 4/118 (3.4) \\
DDR-opt B+ & 9/139 (6.5) & 10/139 (7.2) & 1/118 (0.8) & 3/118 (2.5) \\%
}
\newcommand{\contrastrows}{%
Stop-set & +12.2 & [+6.5, +18.7] & 19/2 \\
Factored, no repair & +4.3 & [+0.0, +8.6] & 8/2 \\
Smac Hybrid B+ & +42.4 & [+33.8, +51.1] & 59/0 \\
Smac Lattice B+ & +34.5 & [+25.2, +43.9] & 55/7 \\
DDR-opt B+ & +37.4 & [+28.1, +46.8] & 57/5 \\%
}

\newcommand{\allrows}{%
Portal-only & 3 & 27 & 10 & 83 & 1 & 29 & 3 & 39 \\
Stop-set & 13 & 44 & 37 & 97 & 5 & 34 & 15 & 49 \\
Stop-set + repair & 11 & 50 & 40 & 98 & 4 & 31 & 16 & 50 \\
Factored, no repair & 22 & 55 & 55 & 102 & 7 & 37 & 19 & 61 \\
GeoDiff & 20 & 61 & 57 & 102 & 6 & 37 & 18 & 59 \\
Smac Hybrid B & 1 & 1 & 2 & 6 & 0 & 0 & 2 & 2 \\
Smac Hybrid B+ & 2 & 2 & 2 & 6 & 0 & 0 & 2 & 2 \\
Smac Lattice B & 2 & 9 & 9 & 25 & 2 & 3 & 4 & 9 \\
Smac Lattice B+ & 2 & 13 & 11 & 26 & 1 & 3 & 4 & 10 \\
DDR-opt B & 2 & 6 & 4 & 10 & 1 & 1 & 2 & 4 \\
DDR-opt B+ & 1 & 9 & 4 & 10 & 1 & 1 & 2 & 3 \\%
}
\begin{document}
\raggedbottom
\begin{frontmatter}
\title{Planning to Stop: Factored Certification and Terminal Repair for Noncircular Robots}

\author[th]{Runtian Zheng}
\author[th]{Congpeng Zhang\corref{cor1}}
\ead{zcp@ncut.edu.cn}
\author[th]{Ying Liu}
\cortext[cor1]{Corresponding author}
\affiliation[th]{organization={North China University of Technology},
            postcode={100000},
            city={Beijing},
            country={China}}

\begin{abstract}
Tight-space navigation requires a robot to clear obstacles and stop inside its goal despite uncertainty. We present GeoDiff, a motion-program planner coupling original-goal stopped-set inclusion, witness-guided terminal repair, and factored proof compilation. An exact feasible-translation characterization guides repair; request-local reuse reduces repeated geometric work. We establish conditions for compiler equivalence and reach-and-stop soundness. Across 11,440 floorplan-derived jobs, GeoDiff certifies 61/139 tasks at 5-cm tolerance within 15 s, versus 44/139 for the stop-set baseline. The strongest observed external certificate adapter certifies 13/139 tasks from 119/139 native routes, separating route availability from certification. On tasks derived from public BARN maps, GeoDiff certifies 202/270 at 5-cm tolerance for a Jackal-shaped body, versus 3/270 for the portal baseline. Development analyses distinguish emitter savings from solver coverage, and 1,420 route-tracking episodes characterize the execution interface. These results show how terminal reasoning and proof reuse increase finite-budget certification coverage while preserving the original physical task and goal.
\end{abstract}

\end{frontmatter}

\section{Introduction}\label{sec:intro}
A robot can negotiate a doorway yet fail to stop inside the space reserved for it. For a noncircular body, a turn changes whole-body clearance; tracking and braking uncertainty spread the possible stopping states. When clearance and terminal tolerance approach the size of those uncertainties, reaching a nominal endpoint is only part of the planning problem. The planner must construct a motion whose complete stopping enclosure fits inside the requested goal.

This requirement also changes how computation should be spent. Route generation, terminal maneuvers, and geometric verification share a finite planning budget. A geometrically plausible route may consume that budget without yielding an acceptable stopping program. Conversely, a candidate that narrowly fails its terminal condition may contain precisely the information needed for an inexpensive repair. The central question is therefore: \emph{how should search use geometric failure evidence and repeated verification work to produce more certified stopping motions under a fixed deadline?}

\begin{figure*}[!tp]
\centering
\includegraphics[width=\textwidth]{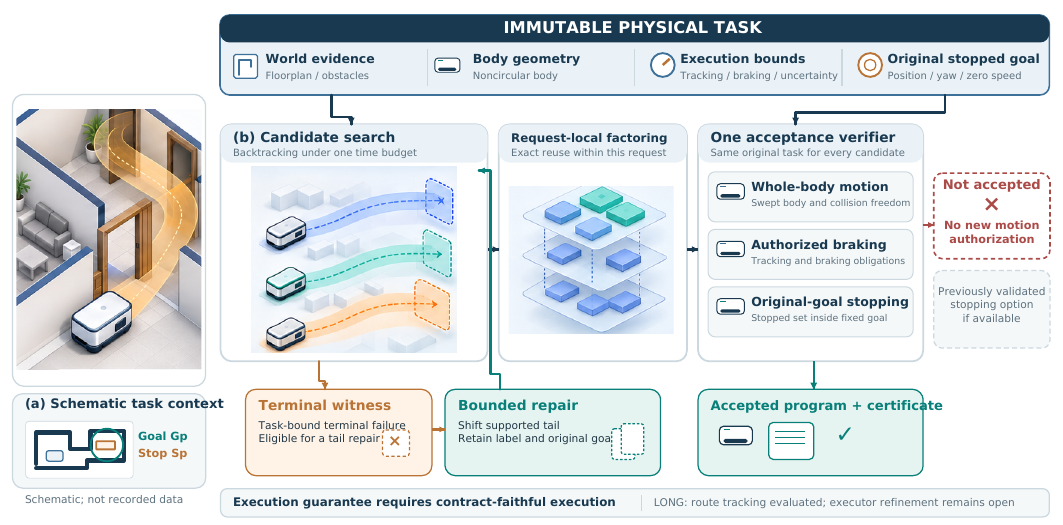}
\caption{GeoDiff organizes search around certification of the immutable original physical task. (a) Conceptual task context with positional projections $G_p$ and $S_p$ of the original goal and stopped enclosure; the scene and route motifs are schematic, not recorded experimental trajectories. (b) Candidate programs pass through exact request-local factoring and one acceptance verifier checking whole-body motion, authorized braking, and original-goal stopping. Task-bound terminal witnesses propose bounded repairs that return to candidate search; neither repair nor factoring changes the original goal or acceptance relation. Physical guarantees additionally require contract-faithful execution.}
\label{fig:method_architecture}
\end{figure*}

We address this question with GeoDiff, a planner organized around a single original-task acceptance relation (Fig.~\ref{fig:method_architecture}). The physical checker establishes whole-body motion and authorized braking obligations. A separate postcondition checks the conservative stopped-state enclosure against the original goal. Eligible terminal failures provide task-bound witnesses for bounded tail adjustments. Every adjusted candidate returns to the same verifier. Request-local factoring reuses identical geometric expressions so that overlapping proof obligations require less repeated computation.

The contribution builds on robust reach--avoid planning, tracking-error bounds, and feedback motion libraries~\cite{parc,neuralparc,fastrack,funnels}. Its distinctive object is the \emph{certifiable motion program under a deadline}: terminal geometry determines which repairs are worth proposing, and proof structure determines which computations can be reused. This coupling connects the stopping specification to both candidate generation and compiler organization.

The paper develops three results. First, it formulates original-goal stopping and derives the exact feasible translation set for a fixed terminal enclosure, including label-dependent repair margins. Second, it establishes the invariants for exact request-local factoring and for transferring accepted finite programs to contract-faithful execution. Third, it evaluates the resulting certification coverage over 11,440 CUBI-DT jobs, reports native-route and certificate endpoints separately, and examines the mechanisms with development analyses and a route-tracking panel. The complete configuration is denoted GeoDiff; \texttt{factored\_bt} is its experiment identifier.

\section{Related Work}\label{sec:related}
\subsection{Whole-Body Geometry and Robust Reach--Avoid}
Implicit swept-volume signed distance fields support continuous collision-free trajectory generation for arbitrary shapes~\cite{svsdf}, and efficient swept-volume planning combines geometric decomposition with trajectory optimization for ground robots~\cite{esv}. Polytope-in-polytope control-barrier formulations address shape-aware containment~\cite{polytope2026}. GeoDiff uses full-body geometry to verify motion and braking, then adds an explicit stopped-set inclusion in the requested goal. Its treatment of common uncertainty follows the dependency-preserving principle of affine arithmetic~\cite{affine_arithmetic}.

PARC computes goal-reaching trajectories near obstacles through piecewise-affine reachability~\cite{parc}; NeuralPARC studies guaranteed reach--avoid for black-box systems through narrow gaps~\cite{neuralparc}. FaSTrack separates planning and tracking using a tracking-error bound~\cite{fastrack}, and funnel libraries compose robust feedback motions~\cite{funnels}. These methods establish the broader foundations for robust goal reaching. Our focus is the finite-budget construction of an acceptable primitive program, specifically the use of terminal witnesses and repeated proof obligations. The experiments compare certificate-producing pipelines, rather than benchmark these reach--avoid systems directly.

\subsection{Certificates, Contracts, and Runtime Assurance}
Proof-carrying code makes evidence independently checkable~\cite{proof_carrying_code}, and Certified Control applies a certificate-and-monitor architecture to autonomous vehicles~\cite{certified_control}. SOTER composes runtime-assurance components~\cite{soter}; VeriPhy studies property preservation through controller compilation~\cite{veriphy}. Platform-aware mission planning uses task-level contracts to feed invalid-plan information back into planning~\cite{platform_contracts2026}. GeoDiff specializes this established separation of proposal and checking to original-goal stopping, with a geometric repair characterization and a compiler organized around reusable obligations.

Control barrier functions constrain admissible controls~\cite{cbf}, while gatekeeper verifies trajectories with a safe backup under stated assumptions~\cite{gatekeeper,gatekeeper2025}. Our guarantee concerns a finite motion program over a declared evidence horizon. Its transfer to execution requires valid enclosures, finite progress, and a controller implementing the certified contract. A covered contingency brake supplies a safe abort; successful arrival additionally requires completion of the designated goal stop.

\subsection{Repair and Compilation}
Temporal behavior tree-guided trajectory repair corrects demonstrations against temporal task specifications before policy learning~\cite{tbt_repair2026}. GeoDiff instead repairs supported terminal primitives using a failed certificate witness. The feasible interval is exact for the fixed positional enclosure, and the rebuilt program undergoes complete original-task verification. Factoring addresses a complementary cost: repeated evaluation of identical geometric expressions. Its equivalence result applies to deterministic completed requests, while its deadline-limited benefit is measured as certification coverage.

\section{Problem Formulation}\label{sec:problem}
\subsection{Task, Motion Program, and Stopped Goal}
Let a task be
\begin{equation}
\T=(X_0,\W,B,\mathcal U,\Xi,G,D,e,H),
\label{eq:task}
\end{equation}
where $X_0$ is an initial-state set; $\W$ is conservative world evidence; $B$ is a body represented by a finite union of convex parts; $\mathcal U$ specifies the execution contract, including input, tracking, latency, and braking bounds; $\Xi$ is a set of shared latent geometric variables; $G$ is the original stopped-goal region; $D$ is the planning budget; and $(e,H)$ identify an evidence epoch and its validity horizon. Disturbance signals belong to $\mathcal D(\mathcal U)$, which is distinct from the scalar budget $D$.

For a ground robot, write $x=(p,\theta,v,\omega)$, where $p\in\R^2$ and $\theta\in S^1$. The body occupancy is $\mathcal O_B(x)=p+R(\theta)B$. A stopped goal may be specified as
\begin{equation}
\begin{split}
G=\{x:\ &\norm{p-p_g}_2\leq\varepsilon_p,\quad
 |\operatorname{wrap}(\theta-\theta_g)|\leq\varepsilon_\theta,\\
&|v|\leq\varepsilon_v,\quad |\omega|\leq\varepsilon_\omega\}.
\end{split}\label{eq:goal}
\end{equation}
The velocity thresholds define the stopped-state convention; exact rest is the special case of zero thresholds. A dwell condition, when used in execution experiments, is an additional temporal requirement and must be evaluated identically for all controllers.

A motion program $\pi$ specifies a finite sequence of supported motion primitives, permitted speed labels, transition conditions, and a terminal or contingency braking policy. Its certificate $C$ carries task binding, interval coverage, geometric separation, tube and input conditions, segment entry/exit relations, and terminal evidence. Optional hints $I$ influence proposal order while the physical task and $G$ remain fixed.

The target acceptance implication is
\begin{equation}
\begin{split}
V(\T,\pi,C)=1\ \Longrightarrow\ 
\forall x_0\in X_0,\ \xi\in\Xi,\ d\in\mathcal D(\mathcal U):\\
\operatorname{SafeMotion}(\pi)\ \wedge\
\operatorname{SafeBrake}(\pi)\ \wedge\ x_{\rm stop}\in G.
\end{split}\label{eq:accept}
\end{equation}
Here the terminal statement concerns execution of the complete accepted program and its designated goal stop. A contingency stop before the goal need only be safe; it is an abort rather than a successful arrival. The same fixed program must cover all contract-admissible uncertainties. In particular, the synthesis quantifiers are $\exists\pi\,\forall(x_0,\xi,d)$, not a separate motion chosen after each disturbance realization.

\subsection{Trust Boundary and Assumptions}
The proposer and compiler supply candidates to a verifier that re-establishes their obligations from the immutable task. A digest binds the certificate to that task and evidence epoch. The conditional guarantees use six assumptions:
\begin{enumerate}
\setlength{\itemsep}{1pt}
\renewcommand{\labelenumi}{A\arabic{enumi}.}
\item Body geometry, primitive semantics, and initial sets enclose the modeled system.
\item Actual tracking, input, delay, and braking behavior lie within $\mathcal U$.
\item World evidence overapproximates occupied space throughout its declared horizon.
\item Shared variables have a common physical origin and coordinate frame; independent residuals retain separate bounds.
\item Outward bounds, interval coverage, and set-inclusion operations in the verifier are sound.
\item Each primitive has validated finite-duration progress from its entry set to its exit enclosure, and its designated brake terminates within a finite bound. Computation, execution, and stopping fit within the evidence horizon. Normal completion proceeds to the designated goal stop without an early contingency abort.
\end{enumerate}
These assumptions define a model-level contract. The experiments below instantiate geometric task models and separately report route-tracking observations. A rejected candidate leaves the task unresolved under the available representation and budget. During execution, abstaining from a new command preserves safety only when the current state has a valid stop under the active contract.

\section{Certificate-Aware Planning}\label{sec:method}
\subsection{One Acceptance Relation, Multiple Proposal Sources}
An adapter instantiates $\T$ from conservative world evidence and a metric goal. Native search proposes supported motions directly; an external adapter converts a route into those primitives. Both use the same acceptance relation $V(\T,\pi,C)$. This common interface makes the certificate endpoint comparable while retaining each proposal source's route-generation behavior.

\subsection{Dependency-Preserving Relative Geometry}\label{sec:relative}
For a fixed outward-bounded motion cell, consider robot and obstacle points represented as
\begin{equation}
 r=r_0+A_r\xi+E_r\eta_r,\qquad
 o=o_0+A_o\xi+E_o\eta_o,
\label{eq:jointrepr}
\end{equation}
where $\xi\in[-1,1]^m$ represents a common physical latent variable, and $\eta_r,\eta_o$ are independent residual variables with bounded sets. Let $n\in S^1$ be a unit separating direction. The support of the relative uncertainty along $n$ is bounded by
\begin{equation}
 \sigma_{\rm joint}(n)=
 \sum_{j=1}^m|n^\top(a_{r,j}-a_{o,j})|+\rho_r(n)+\rho_o(-n),
\label{eq:support}
\end{equation}
with $\rho_r,\rho_o$ the residual support bounds. Replacing the common latent by independent copies yields
\begin{equation}
 \sigma_{\rm marg}(n)=
 \sum_j\bigl(|n^\top a_{r,j}|+|n^\top a_{o,j}|\bigr)
 +\rho_r(n)+\rho_o(-n).
\label{eq:marg}
\end{equation}

\begin{lemma}[Relative support bound]\label{lem:relative}
Under the affine representation in~\eqref{eq:jointrepr}, $\sigma_{\rm joint}(n)\leq\sigma_{\rm marg}(n)$. A truly common translation has zero contribution to the relative uncertainty.
\end{lemma}
\begin{proof}
The support of a linear image of a box is the sum of absolute projected generator coefficients. Subtracting the two representations before maximizing gives~\eqref{eq:support}. For each generator, $|n^\top(a_{r,j}-a_{o,j})|\leq|n^\top a_{r,j}|+|n^\top a_{o,j}|$. Summation proves the inequality. If the columns are the same realization of a common translation, $a_{r,j}=a_{o,j}$, so their difference vanishes.
\end{proof}

The representation binds shared columns to the same physical latent variable. For finite yaw intervals, a validated pull-back and an outward remainder enclose the nonlinear rotation. Independent residuals remain in $\rho_r$ and $\rho_o$; numerical equality of columns alone does not establish a shared error source.

For robot part $B_a$ and obstacle $O_b$, a sufficient cell-wise separation certificate chooses $n$ such that
\begin{equation}
 \inf_{z\in\widehat{B_a-O_b}}n^\top z\ \geq\ m_{ab}>0,
\label{eq:sep}
\end{equation}
where $\widehat{B_a-O_b}$ encloses relative occupancy over the entire cell, with shared variables preserved before maximization. The normal can differ across cells and body/obstacle pairs. The construction also applies to boundary containment and admissible braking occupancy. The strictly positive margin excludes contact; a zero margin establishes only nonpenetration.

\begin{figure}[!t]
\centering\includegraphics[width=\columnwidth]{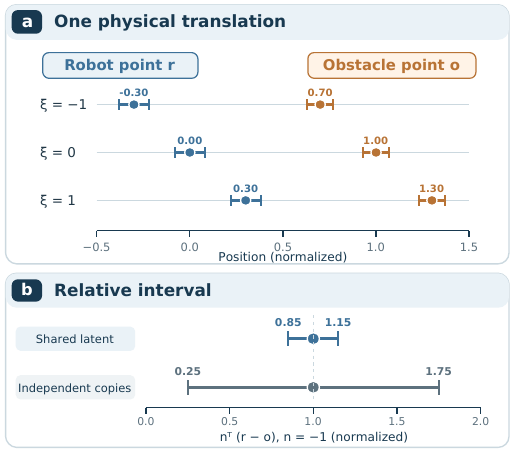}
\caption{Shared-variable relative geometry in a one-dimensional normalized example: $r=0.3\xi+e_r$, $o=1+0.3\xi+e_o$, $|\xi|\leq1$, $|e_r|\leq0.08$, and $|e_o|\leq0.07$. (a) The same translation moves both points, while residuals remain independent. (b) Projecting $r-o$ along $n=-1$ gives $[0.85,1.15]$; replacing the shared variable by independent copies gives $[0.25,1.75]$.}
\label{fig:shared_uncertainty}
\end{figure}

Figure~\ref{fig:shared_uncertainty} illustrates how relative geometry retains the dependency.

\subsection{Physical Obligations and Original-Goal Stopping}\label{sec:stop}
A terminal pose check can be unnecessarily restrictive when it couples a legacy acceptance representation to the goal. We separate two objects: $\Phys(\T,\pi,C)$, containing the physical swept-body, braking, tube, input, and composition obligations; and a conservative stopped-state enclosure $S_{\rm stop}(\T,\pi,C)$. The target condition is
\begin{equation}
 \Phys(\T,\pi,C)\quad\wedge\quad S_{\rm stop}(\T,\pi,C)\subseteq G.
\label{eq:decompose}
\end{equation}

An auxiliary task $\T_{\rm aux}$ permits reuse of the physical verifier by replacing only a legacy terminal tolerance or identifier. It preserves $X_0,\W,B,\mathcal U,\Xi,e,H$. Final acceptance additionally replays the stopped set under the \emph{original} task and checks its position, yaw, and stopped-state requirements.

\begin{proposition}[Original-goal postcondition decomposition]\label{prop:goal}
Suppose acceptance on $\T_{\rm aux}$ implies $\Phys(\T,\pi,C)$, and every admissible terminal state of the complete program lies in $S_{\rm stop}(\T,\pi,C)$. If $S_{\rm stop}(\T,\pi,C)\subseteq G$, then the original physical obligations and stopped-goal requirement hold.
\end{proposition}
\begin{proof}
The first premise transfers the physical obligations from the auxiliary check without changing their task parameters. For any admissible execution, the enclosure premise gives $x_{\rm stop}\in S_{\rm stop}(\T,\pi,C)$. The inclusion in $G$ yields $x_{\rm stop}\in G$ by transitivity. Both conclusions refer to the original task.
\end{proof}

The decomposition exposes original-goal inclusion as a replayable obligation and reuses the physical checker. Each supported tail type requires a valid terminal enclosure. This change makes the stopping postcondition explicit; comparisons with the portal-only predecessor therefore measure representation and search together.

\subsection{Witness-Guided Terminal Repair}\label{sec:repair}
When a candidate passes enough physical checks to yield a failed terminal replay, the compiler records a witness containing the task digest, tail identity, speed label, stopped enclosure, failure reason, and translation eligibility. Ordinary template backtracking remains the first search mechanism. Only eligible terminal witnesses enter a bounded repair queue; unrelated failures do not justify a terminal shift.

For a fixed supported tail and fixed speed label, write the positional stopped-set enclosure relative to the original goal as
\begin{equation}
 S_p(\delta)=\conv\{v_i+\delta d:i=1,\ldots,k\}\oplus r\B_2,
\label{eq:repairset}
\end{equation}
where $k\geq1$, $d\in\R^2$ is a permitted translation direction, $\delta$ is its scalar magnitude, $r\geq0$ is a conservative residual radius, and $\B_2$ is the unit Euclidean disk. The positional tolerance is $\varepsilon_p\geq0$, and permitted shifts satisfy $|\delta|\leq\delta_{\max}$.

\begin{proposition}[Feasible interval for a fixed terminal enclosure]\label{prop:repair}
For~\eqref{eq:repairset}, inclusion $S_p(\delta)\subseteq\varepsilon_p\B_2$ is equivalent to $r\leq\varepsilon_p$ and
\begin{equation}
 a\delta^2+2b_i\delta+c_i\leq0\quad\text{for all }i,
\label{eq:quad}
\end{equation}
where $a=\norm{d}_2^2$, $b_i=v_i^\top d$, and $c_i=\norm{v_i}_2^2-(\varepsilon_p-r)^2$. Intersecting these sublevel sets with $[-\delta_{\max},\delta_{\max}]$ gives a closed interval, a point, or the empty set.
\end{proposition}
\begin{proof}
Let $K=\conv\{v_i+\delta d\}$. For a nonempty compact $K$, $\max_{z\in K\oplus r\B_2}\norm z_2=\max_{z\in K}\norm z_2+r$: the upper bound follows from the triangle inequality, and equality is achieved by adding a radius-$r$ vector in the direction of a norm-maximizing point (with the zero case immediate). By convexity, $\max_{z\in K}\norm z_2=\max_i\norm{v_i+\delta d}_2$. Thus the inclusion is equivalent to all vertex norms being at most $\varepsilon_p-r$. Squaring yields~\eqref{eq:quad}. Each inequality is a convex quadratic sublevel set; their intersection with a closed bounded interval has the stated form.
\end{proof}

\begin{figure*}[!tp]
\centering\includegraphics[width=\textwidth]{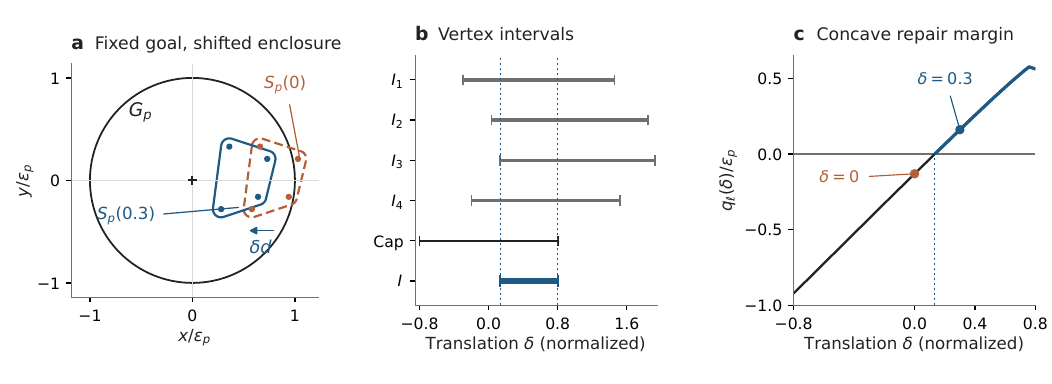}
\caption{Analytical fixed-enclosure repair example, with normalized $\varepsilon_p=1$, $r=0.08$, $d=(-1,0)$, and $|\delta|\leq0.8$. (a) Original and translated stopping enclosures relative to the fixed positional goal $G_p=\varepsilon_p\B_2$. (b) The vertex-wise quadratic feasible intervals intersect the translation cap in $I=[0.1343,0.8]$. (c) The concave margin $q_\ell(\delta)$ admits the illustrated shift $\delta=0.3$. The rebuilt program also requires complete verification of yaw, dynamics, and swept motion.}
\label{fig:repair_geometry}
\end{figure*}

\begin{corollary}[Label-dependent repair margin]\label{cor:margin}
For a fixed label $\ell$ with enclosure vertices $v_{i,\ell}$, residual $r_\ell$, and direction $d_\ell$, define
\begin{equation}
q_\ell(\delta)=\varepsilon_p-r_\ell-
\max_i\norm{v_{i,\ell}+\delta d_\ell}_2.
\end{equation}
Then $q_\ell$ is concave. Its nonnegative superlevel set intersected with the permitted translation interval is exactly the positional feasible set in Proposition~\ref{prop:repair}. Feasibility over a finite label family is the union of these label-specific sets and need not be one interval.
\end{corollary}
\begin{proof}
Each norm is convex in $\delta$, and a pointwise maximum of convex functions is convex. Negation gives concavity; the superlevel statement follows from the same vertex inclusion condition. A union of intervals can be disconnected, proving the final assertion.
\end{proof}
The margin gives repair a geometric objective: translation improves terminal inclusion within a label, while a label change can alter the enclosure itself (Fig.~\ref{fig:repair_geometry}). The feasible set provides candidate shifts for the search queue.

For $a>0$ and $b_i^2-ac_i\geq0$, the $i$th interval endpoints are
\begin{equation}
 \ell_i=\frac{-b_i-\sqrt{b_i^2-ac_i}}{a},\qquad
 u_i=\frac{-b_i+\sqrt{b_i^2-ac_i}}{a}.
\label{eq:roots}
\end{equation}
A negative discriminant rejects that fixed enclosure. If $d=0$, the condition reduces to $r\leq\varepsilon_p$ and $c_i\leq0$ for all $i$. Numerical proposal generation may use approximations to these endpoints, but the final verifier must conservatively recheck the resulting program.

A translated tail changes swept occupancy, and recompilation may change its stopping enclosure. The fixed-enclosure characterization therefore serves proposal generation: every repair is rebuilt and checked for swept motion, yaw, dynamics, and original-goal stopping under $V$. The requested destination remains fixed.

\begin{figure*}[!tp]
\centering
\includegraphics[width=\textwidth]{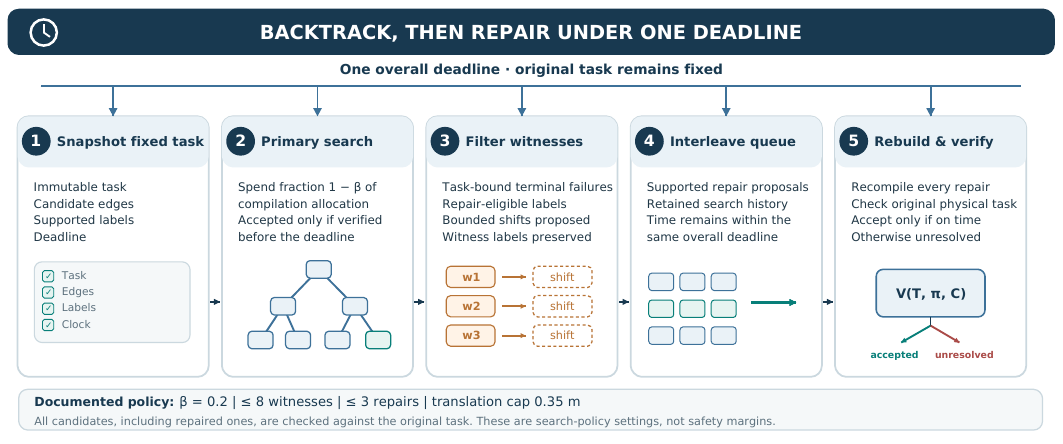}
\caption{One-deadline backtracking and witness-guided terminal repair policy. The immutable task, supported labels, search history, and deadline are retained throughout. Primary search receives fraction $1-\beta$ of the compilation allocation; eligible terminal witnesses propose label-preserving bounded repairs, which are rebuilt and checked against the original task. Only accepted, on-time programs are returned. The documented configuration uses $\beta=0.2$, at most eight witnesses, at most three repairs, and a 0.35 m translation cap; these are search-policy settings, not safety margins.}
\label{alg:repair}
\end{figure*}

The policy in Fig.~\ref{alg:repair} reserves a fraction of the allocation for repair while retaining unused primary time. Its empirical value depends on the deadline and the candidate distribution; Section~\ref{sec:results} reports both gains and reversals.

\begin{figure*}[!tp]
\centering\includegraphics[width=\textwidth]{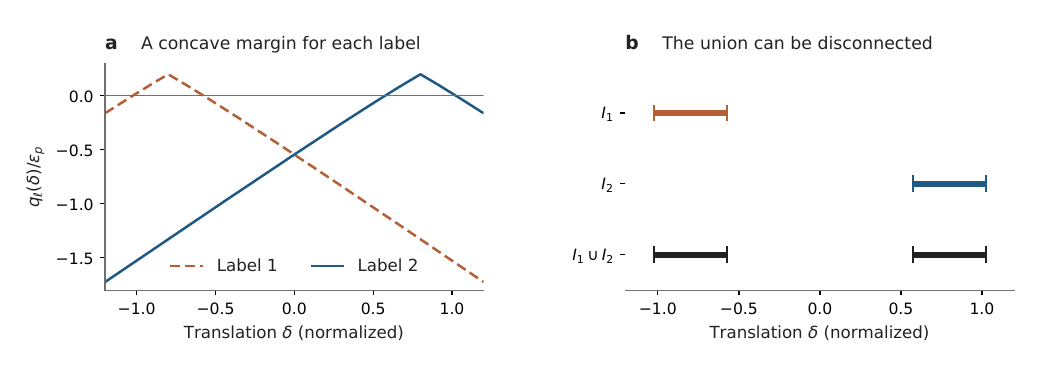}
\caption{Analytical illustration of label-dependent feasibility. Two rectangular vertex sets, centered at $(\pm0.8,0)$ with half-widths $(0.55,0.35)$, use $\varepsilon_p=1$, $r=0.15$, $d=(-1,0)$, and $|\delta|\leq1.2$. (a) Each margin is concave. (b) The feasible intervals are approximately $[-1.0246,-0.5754]$ and $[0.5754,1.0246]$, whose union is disconnected. These specified enclosures illustrate Corollary~\ref{cor:margin}.}
\label{fig:label_feasibility}
\end{figure*}

The label-wise characterization also explains why repair should retain the label attached to a witness. Each fixed label has an interval of feasible translations; the union across labels can be disconnected (Fig.~\ref{fig:label_feasibility}).

\subsection{Exact Factoring of Proof Obligations}\label{sec:factoring}
Overlapping candidates repeatedly evaluate the same motion-cell rotation bounds, directional body supports, and wall separations. The factored compiler represents these computations as an expression directed acyclic graph and reuses repeated occupancy-cell, rotation, support, body-part, obstacle-separation, and wall-bound expressions within one immutable request.

The required cache invariant is that a key contains all arguments on which the stored expression depends, including interval endpoints, primitive and part identity, separating normal, uncertainty representation, braking/stopping modes, and immutable task/epoch binding. Each request must be snapshotted before compilation. A bounded least-recently-used table may evict entries, but reuse through an incomplete key is disallowed. The acceptance kernel must recheck the obligations independently of the compiler's cache.

\begin{proposition}[Completed-request equivalence]\label{prop:cache}
Suppose factored and unfactored compilers receive the same immutable request, use identical outward arithmetic and deterministic traversal/selection rules, and both complete. If every cache key identifies the complete inputs of its expression, they produce identical intermediate values and the same deterministic certificate serialization.
\end{proposition}
\begin{proof}
Order the expression graph topologically. Leaves agree because the request and arithmetic inputs agree. Assume all predecessors of a node agree. A miss evaluates the same function of equal predecessors; a hit returns a previously evaluated value for those exact inputs. The node values therefore agree. Induction establishes agreement throughout the graph. Deterministic traversal and serialization then give the same output. Eviction changes only whether an equal value is recomputed.
\end{proof}

For $Q$ requests to an expensive expression and $U$ distinct keys without eviction, work changes from approximately $Q c_f$ to $U c_f+Q c_\ell$, where $c_f$ and $c_\ell$ are evaluation and lookup costs. Reuse is advantageous when saved evaluations outweigh lookup overhead. With $U=Q$, or a dominant cost elsewhere, the saving can vanish. Under a deadline, altered completion times also change which later candidates are reached. The emitter and full-task experiments therefore measure equivalence, local cost, and solver coverage separately.

\section{Conditional Safety and Composition}\label{sec:theory}
\subsection{Continuous Program Soundness}
Let segment $j$ have a validated entry set $E_j$, an exit enclosure $F_j$, and a finite collection of closed cells covering its entire parameter/time domain. The cell evidence must enclose ordinary occupancy and braking occupancy from every trigger state authorized by the contract. Checking only the nominal terminal brake does not cover arbitrary mid-segment intervention.

\begin{lemma}[Finite physical composition]\label{lem:composition}
Under A1--A5 and the finite-duration progress premise of A6, suppose every segment's cells completely cover its domain and verify its tube, input, full-body separation, boundary, and authorized braking obligations. If $X_0\subseteq E_1$ and $F_j\subseteq E_{j+1}$ at every transition, the finite program satisfies $\Phys(\T,\pi,C)$ throughout its valid horizon.
\end{lemma}
\begin{proof}
The entry inclusion and the tube conditions keep each admissible execution of the first segment in the enclosure covered by its cells. Each cell's outward separation bounds apply to every body part at all times in that cell, including the covered braking continuations. Complete coverage prevents an unchecked interval between cells. At the transition, the exit enclosure lies in the next validated entry set. Induction over the finite segment sequence proves the same properties for all segments. The boundary and input conditions are inherited at each step. The finite-duration exit and braking bounds in A6 ensure normal completion and a designated terminal stop in finite time.
\end{proof}

\begin{theorem}[Conditional reach-and-stop soundness]\label{thm:main}
Assume A1--A6. Let $V$ check immutable task/epoch binding, all hypotheses of Lemma~\ref{lem:composition}, and the hypotheses of Proposition~\ref{prop:goal} for the designated terminal stop. An accepted $(\pi,C)$ satisfies~\eqref{eq:accept} for normal completion over the declared horizon. A covered early contingency brake preserves safety but is an abort and need not reach $G$. This conclusion is unchanged if the candidate was obtained by bounded terminal repair or exact factored compilation.
\end{theorem}
\begin{proof}
Lemma~\ref{lem:composition} establishes all physical and authorized braking obligations for every admissible execution. The auxiliary task preserves those physical fields, and terminal replay encloses every designated stopped state. Proposition~\ref{prop:goal} places those states in the original $G$. Task and epoch binding ensure that these are the obligations for the task being executed, and A6 ensures their temporal applicability. Repair and factoring affect only the candidate supplied to $V$; acceptance re-establishes the same premises, so neither transformation changes the conclusion.
\end{proof}

The theorem identifies the proof obligations connecting acceptance to execution. The implementation evidence consists of certificate replay and geometric audits; it does not include a machine-checked kernel proof. A different controller must be shown to respect $\mathcal U$ before the conclusion transfers to centerline tracking.

\subsection{Source Independence and Local Continuation}
\begin{proposition}[Source-independent acceptance]\label{prop:source}
If two adapters produce the same physical task $\T$, source hints affect only candidate generation, and $V$ has no dependence on the source label, then any fixed $(\pi,C)$ receives the same verdict under both adapters.
\end{proposition}
\begin{proof}
The verifier receives identical arguments and immutable task fields. Its verdict is therefore identical. Differences in proposal order may change whether a candidate is found by a deadline, but not the acceptance of that fixed candidate.
\end{proof}

Source independence preserves the verdict for a fixed physical task and candidate. Changes in world evidence, initial sets, or uncertainty define a different $\T$ and require a new check.

\begin{corollary}[Finite local continuation]\label{cor:local}
Consider an existing finite sequence of accepted programs satisfying Theorem~\ref{thm:main}. Suppose each predecessor's complete stopped-state enclosure $S_j$ lies in the next initial set $X_{0,j+1}$, and each successor's evidence remains valid through its execution and authorized stop. Their concatenation is safe for that finite sequence. If the execution contract has controller modes, queued commands, or delay states, the handoff sets must enclose those augmented states as well.
\end{corollary}
\begin{proof}
The first program terminates within $S_1$. Its inclusion in $X_{0,2}$ establishes the next program's initial premise. Repeated application of Theorem~\ref{thm:main} proves the result by finite induction. Each temporal premise preserves the applicability of the corresponding world evidence.
\end{proof}

During local continuation, the supervisor must preserve sufficient time and distance for an already valid stop while a successor is pending. Each handoff requires fresh applicable evidence. The corollary covers a finite accepted sequence; it assumes neither successor existence nor indefinite validity of a stopped state in a changing world.

\section{Experimental Design}\label{sec:experiments}
\subsection{Door-and-Turn Tasks Derived from Floorplans}
We derive CUBI-DT door-and-turn tasks from the semantic floorplan geometry of CubiCasa5K~\cite{cubic}. Metric scale is assigned by normalizing each floorplan's median annotated door major-axis length to 0.9 m or 0.8 m. These values are geometric assumptions applied to the floorplan, rather than measured widths of every selected opening.

A deterministic source-hash rule selects the door. The task uses an $8\times8$ m crop, a $0.90\times0.62$ m rectangular body, and a start 1.5 m before the door along its normal. A goal is selected from a 0.1 m grid in the destination room, between 1.5 and 2.5 m from the door, with at least 0.8 m lateral displacement and 0.6 m obstacle clearance. The goal yaw differs by $\pm\pi/2$ from the door normal. These rules require entry followed by a turn, rather than an almost straight doorway crossing. Walls with door openings, fixed furniture, columns, stairs, and unannotated exterior space determine the derived obstacles. Unsupported geometry is rejected; polygon repairs are recorded.

The resulting 260 tasks span 140 floorplan clusters: 257 construction-positive tasks with admissible endpoints and three blocked controls. The 0.9 m and 0.8 m panels contain 139 and 118 positive tasks, respectively. Construction positivity supplies the evaluation denominator; it is not an oracle of robust feasibility. Job rows record the body dimensions and a 0.01 m independent map-residual bound. Appendix~\ref{app:repro} identifies the retained parameters and the task-contract fields needed for full reconstruction.

The distribution deliberately concentrates on turning and stopping difficulty. The frozen protocol excludes 80 development floorplan clusters and uses a recorded configuration with staged 10\%/90\% inspection. We therefore report descriptive statistics and retain floorplan variants within their source cluster when estimating uncertainty.

\subsection{Methods and Conversion Contract}\label{sec:baselines}
The primary experiment includes five internal settings and six external settings. Portal-only is the earlier portal-based acceptance/search configuration. Stop-set introduces the explicit original-goal stopping condition. Stop-set + repair adds witness-guided repair without factored compilation. Factored, no repair isolates the compiler change relative to Stop-set. GeoDiff combines stop-set acceptance, factoring, and repair. The four stop-set settings form a $2\times2$ factorial comparison of factoring and repair; Portal-only additionally changes acceptance representation.

External candidates come from Smac Hybrid-A*, Smac State Lattice~\cite{smac}, and DDR-opt~\cite{ddropt}. Each has two postprocessing settings: B uses factored compilation without repair; B+ also permits repair. Both are checked by the same kernel against the original task. They are adapter configurations, not the authors' native systems endowed with our certificate format.

The frozen v2 converter reserves a 0.25 m straight terminal segment for the supported stopping primitive. It trims the native tail, constructs an approach point along the original goal heading and travel direction, converts the remaining route, and appends a zero-curvature tail. Goal and tolerance remain unchanged, although the converted motion can differ from the native trajectory. An adapter failure therefore concerns the complete conversion-and-certification pipeline. All six settings appear in the tables, with the strongest \emph{observed} certificate adapter used as a descriptive reference. Native-route availability is reported separately.

\subsection{Budget, Metrics, and Recount}
The Cartesian product of 260 tasks, 11 settings, two position tolerances (0.05/0.15 m), and two budgets (5/15 s) produces 11,440 jobs. The primary endpoint is strict certified success: an original-task accepted certificate returned within the online budget. A late certificate remains a failure even if a later audit accepts it. Success rates use every positive task in the condition as the denominator, rather than only representable or successfully converted routes.

External timing includes native planning, conversion, compilation/repair, and online verification. Internal timing covers the binding run and child startup; worker imports are excluded. External producer startup is amortized outside that stream. This process-boundary asymmetry makes the measurements specific to the recorded run. Offline replay is reported separately. TEST uses one worker; the recorded environment is Linux 6.8, Python 3.10.12, NumPy 2.2.6, SciPy 1.15.3, and Shapely 2.1.2. Complete campaign-specific CPU binding is unavailable.

Counts, plots, and paired differences are recomputed from the retained JSONL records. A paired bootstrap resamples floorplan clusters 20,000 times, retaining all method outcomes within each sampled cluster. Each single-normalization panel has one positive task per included floorplan. We report pointwise descriptive percentile 95\% intervals for rate differences, without multiplicity adjustment. This analysis reuses recorded outcomes.

\subsection{Evidence Roles and Execution Panel}
The sealed CUBI-DT TEST campaign supplies the primary certification comparison. A separate public-map panel extends the geometric setting; LONG measures route-tracking outcomes. P1 contains development analyses of the emitter, full-task builder, and repair policies, including reruns of exposed tasks. P0 evaluates the sensitivity of LONG to 15 s source-certificate selection. These evidence sets retain their distinct endpoints and denominators.

LONG compares certificate-derived route tracking with Smac Lattice followed by MPPI and Collision Monitor~\cite{smac,mppi,collisionmonitor}. It contains 28 CUBI tasks and 43 BARN-derived tasks~\cite{barn}, each with five seeds in clean and disturbed settings for both systems: 1,420 episodes in total. The 28 CUBI tasks span 27 floorplan clusters. Seeds are repeated trials within tasks, not independent new maps. BARN scenes were previously exposed and are not presented as an untouched generalization split.

LONG selects the longest-budget available accepted certificate. Among 18 CUBI tasks with certificates, source budgets are 15 s for nine tasks, 60 s for four, 120 s for four, and 300 s for one. It therefore evaluates tracking from available certificates rather than matched online planning. The strict endpoint checks task-based position, yaw, linear/angular stopping thresholds, and dwell. The interface endpoint additionally accepts native action success. Section~\ref{sec:trackc} describes the observation differences affecting those endpoints.

\section{Results}\label{sec:results}
\subsection{Finite-Budget Certification Coverage}
Table~\ref{tab:primary} gives all configurations at 15 s. GeoDiff certifies 61/139 tasks (43.9\%) at the 0.9 m normalization and 0.05 m tolerance, compared with 44/139 (31.7\%) for Stop-set. The paired improvement is 12.2 percentage points, with a descriptive 95\% interval of [6.5, 18.7]. Against Smac Lattice B+, the strongest observed external setting in that condition, the difference is 34.5 points [25.2, 43.9]. The latter comparison concerns the end-to-end certificate adapter, including representation conversion.

\paragraph{Separating route availability from certification.}
Table~\ref{tab:external_endpoints} exposes the gap between finding a route and certifying a stopping program. Lattice B+ supplies 119/139 native routes and 13/139 on-time accepted certificates; Hybrid B+ yields 124/139 and 2/139, and DDR B+ yields 111/139 and 9/139. Route availability remains high while the converted program must satisfy the stricter certificate obligations. The certificate column measures the complete adapter pipeline, rather than native navigation success.

\begin{table}[!t]
\caption{External baseline endpoints in the primary CUBI-DT condition (139 positive tasks; 15 s). Native path and accepted certificate are different endpoints.}
\label{tab:external_endpoints}\centering\small
\begin{tabular}{lrr}\toprule
External setting & Native path & Accepted certificate\\\midrule
Smac Lattice B+ & 119/139 & 13/139\\
Smac Lattice B & 119/139 & 9/139\\
Hybrid B+ & 124/139 & 2/139\\
Hybrid B & 124/139 & 1/139\\
DDR B+ & 111/139 & 9/139\\
DDR B & 111/139 & 6/139\\\bottomrule
\end{tabular}
\end{table}

\begin{table*}[!tp]
\caption{Certified success within 15 s: count/positive tasks (percentage). Every listed method is evaluated on the same task set within a column. B and B+ denote external conversion followed by factored compilation without and with repair, respectively.}
\label{tab:primary}\centering\small
\begin{tabular}{lrrrr}\toprule
&\multicolumn{2}{c}{0.9 m door-median normalization}&\multicolumn{2}{c}{0.8 m door-median normalization}\\
\cmidrule(lr){2-3}\cmidrule(lr){4-5}
Method & 0.05 m goal & 0.15 m goal & 0.05 m goal & 0.15 m goal\\\midrule
\primaryrows
\bottomrule
\end{tabular}
\end{table*}

\begin{figure*}[!tp]
\centering\includegraphics[width=\textwidth]{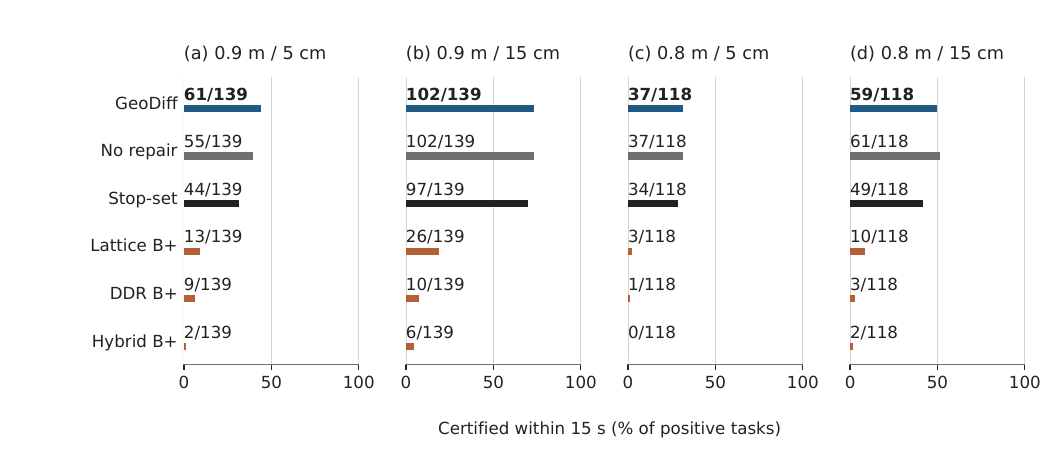}
\caption{Finite-budget certification coverage across the four 15 s conditions. The complete table includes all external settings; the plot shows a subset for readability. The normalizations describe assumed floorplan scale, not a measured door-width sweep on one fixed building.}
\label{fig:success}
\end{figure*}

At 0.15 m tolerance, GeoDiff reaches 102/139 (73.4\%) in the 0.9 m panel. In the 0.8 m panel it reaches 37/118 (31.4\%) at 0.05 m and 59/118 (50.0\%) at 0.15 m. Figure~\ref{fig:success} separates these task distributions: normalization changes both geometry and the retained task set.

\begin{figure*}[!tp]
\centering\includegraphics[width=\textwidth]{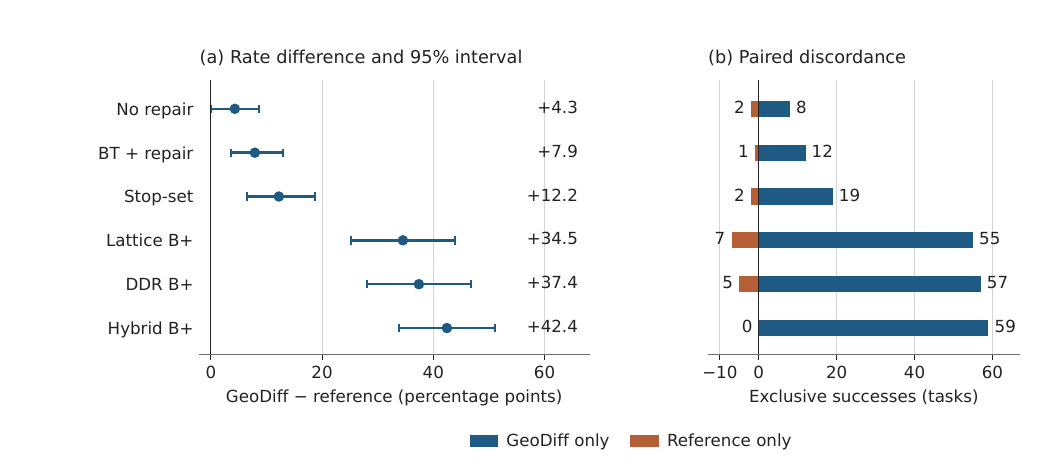}
\caption{Paired outcomes in the primary TEST condition (0.9 m normalization, 0.05 m tolerance, 15 s; 139 tasks). (a) GeoDiff-minus-reference differences with descriptive, pointwise 95\% floorplan-cluster bootstrap intervals. (b) Exclusive successes in each direction. The difference against the no-repair configuration is positive while its interval touches zero; reverse exclusive successes remain visible.}
\label{fig:paired}
\end{figure*}

The paired outcomes explain the aggregate gain (Table~\ref{tab:contrasts}, Fig.~\ref{fig:paired}). GeoDiff succeeds on 19 primary tasks where Stop-set fails and loses on two; the corresponding counts against Lattice B+ are 55 and seven. Finite-budget search thus exchanges some task outcomes while increasing total coverage.

\begin{table}[!t]
\caption{Paired differences for the 0.9 m, 0.05 m, 15 s condition. Differences and pointwise 95\% intervals are in percentage points. Discordance is GeoDiff-only/reference-only success.}
\label{tab:contrasts}\centering\scriptsize
\begin{tabular}{lrrr}\toprule
Reference & Difference & 95\% interval & Discordance\\\midrule
\contrastrows
\bottomrule\end{tabular}
\end{table}

\subsection{Certification Tasks from Public BARN Maps}
\begin{figure*}[!tp]
\centering\includegraphics[width=\textwidth]{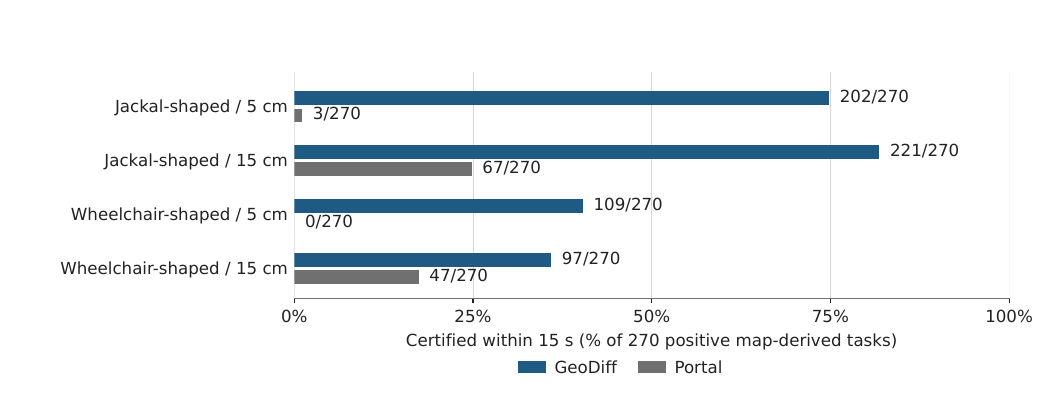}
\caption{Certification coverage on tasks constructed from public BARN maps. All four body/tolerance conditions use 270 positive tasks and a 15 s budget. The plot compares GeoDiff with the portal-based predecessor under the recorded certification contract.}
\label{fig:barn_certification}
\end{figure*}

A separate panel constructs stopping tasks from 300 public BARN static maps using fixed start/goal pairs and the same uncertainty and braking contract. Thirty maps are assigned to development and 270 to the positive TEST panel; added-wall negative variants yield zero certificates. Table~\ref{tab:barn_supplement} and Fig.~\ref{fig:barn_certification} report all four body/tolerance conditions at 15 s. For the Jackal-shaped body, GeoDiff certifies 202/270 tasks at 0.05 m and 221/270 at 0.15 m, versus 3/270 and 67/270 for Portal-only. For the wheelchair-shaped body, the corresponding counts are 109/270 and 97/270, versus 0/270 and 47/270. The lower wheelchair count at the looser tolerance illustrates the nonmonotone behavior of budget-limited search. These are constructed certification tasks, distinct from official BARN navigation scores.

The subsequent replay audit lists 841 records across the campaign: 829 available certificates are accepted by V08, and 12 records have no certificate available. Replay coverage therefore applies to those 829 certificates; the primary online counts retain their recorded endpoint.

\begin{table*}[!tp]
\caption{All body/tolerance conditions in the public-BARN-map-derived certification panel. Counts are on-time certificates / 270 positive tasks at a 15 s budget.}
\label{tab:barn_supplement}\centering\small
\begin{tabular}{llrr}\toprule
Body model & Goal tolerance & \texttt{factored\_bt} & \texttt{pcg\_portal}\\\midrule
Jackal-shaped & 0.05 m & 202/270 (74.8\%) & 3/270 (1.1\%)\\
Jackal-shaped & 0.15 m & 221/270 (81.9\%) & 67/270 (24.8\%)\\
Wheelchair-shaped & 0.05 m & 109/270 (40.4\%) & 0/270 (0.0\%)\\
Wheelchair-shaped & 0.15 m & 97/270 (35.9\%) & 47/270 (17.4\%)\\\bottomrule
\end{tabular}
\end{table*}

\subsection{Mechanisms and Deadline Dependence}
The stop-set $2\times2$ comparison separates the available component settings. At 0.9 m/0.05 m/15 s, Stop-set certifies 44 tasks, repair alone 50, factoring alone 55, and their combination 61. The gains of six, eleven, and seventeen tasks indicate complementary benefits in this condition.

The repair allocation has a measurable tradeoff. Both factored variants certify 102 tasks at 0.9 m/0.15 m and 37 at 0.8 m/0.05 m. At 0.8 m/0.15 m, factoring alone reaches 61 while GeoDiff reaches 59. Time spent repairing terminal failures can improve some tasks while leaving less time for other search branches.

The deadline changes that tradeoff: at 5 s, GeoDiff certifies 20/139 primary tasks versus 22/139 for factoring alone; at 15 s the ordering becomes 61 versus 55. Appendix~\ref{app:counts} reports the complete budget/tolerance matrix.

\begin{figure*}[!tp]
\centering\includegraphics[width=\textwidth]{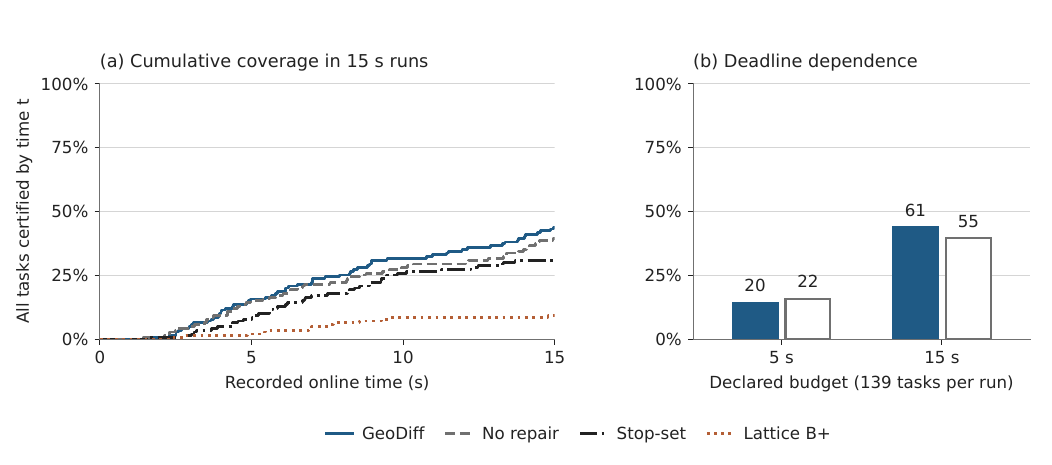}
\caption{Deadline dependence in the primary TEST condition. (a) Unconditional cumulative coverage from the recorded 15 s runs, with all 139 positive tasks retained in each denominator. (b) Separate 5 s and 15 s runs of GeoDiff and the no-repair configuration; bar labels give certified task counts. Recorded online time follows the process boundaries in Section~\ref{sec:experiments}.}
\label{fig:testtime}
\end{figure*}

Figure~\ref{fig:testtime} shows unconditional time-to-certificate curves for the 15 s run, with all 139 positive tasks in the denominator. Their interpretation follows the recorded timing convention in Section~\ref{sec:experiments}. A fresh 5 s run need not follow the same search prefix because repair allocation depends on the declared budget.

\subsection{Development Mechanisms: Where Coverage Is Gained}\label{sec:p1}
P1 examines representation loss, emitter cost, and host repair behavior through development analyses. Its endpoint is reverified acceptance under the stated host configuration. Repetitions are grouped by task, and their outcomes remain separate from the sealed TEST endpoint.

\textit{E1: loss after route conversion.} Native-found, convertible, and strictly certified counts are 119/119/13 for Lattice B+, 124/124/2 for Hybrid B+, and 111/111/9 for DDR B+. Thus substantial attrition occurs after conversion. Geometric and brake-related intermediate indicators are inferred, non-nested fields; they localize a downstream certification bottleneck without assigning a measured fraction to braking alone.

\textit{E2: proof reuse improves the emitter and solver in different ways.} All 195 paired emitter requests from 65 fresh, success-biased development fixtures pass the recorded equivalence checks. The median unfactored-to-factored emitter time ratio is 1.609. A separate intervention on 139 full tasks with three repeats yields 177/417 reverified acceptances with factoring and 88/417 without. At least one repeat accepts on 60/139 versus 32/139 tasks: 28 are factored-only, with none in the reverse direction; majority acceptance is 59 versus 29. The median full-solver time ratio is 1.010. Together these measurements show reduced emitter work and higher task coverage, with the timing ratio specific to the measured component.

\begin{figure*}[!tp]
\centering\includegraphics[width=\textwidth]{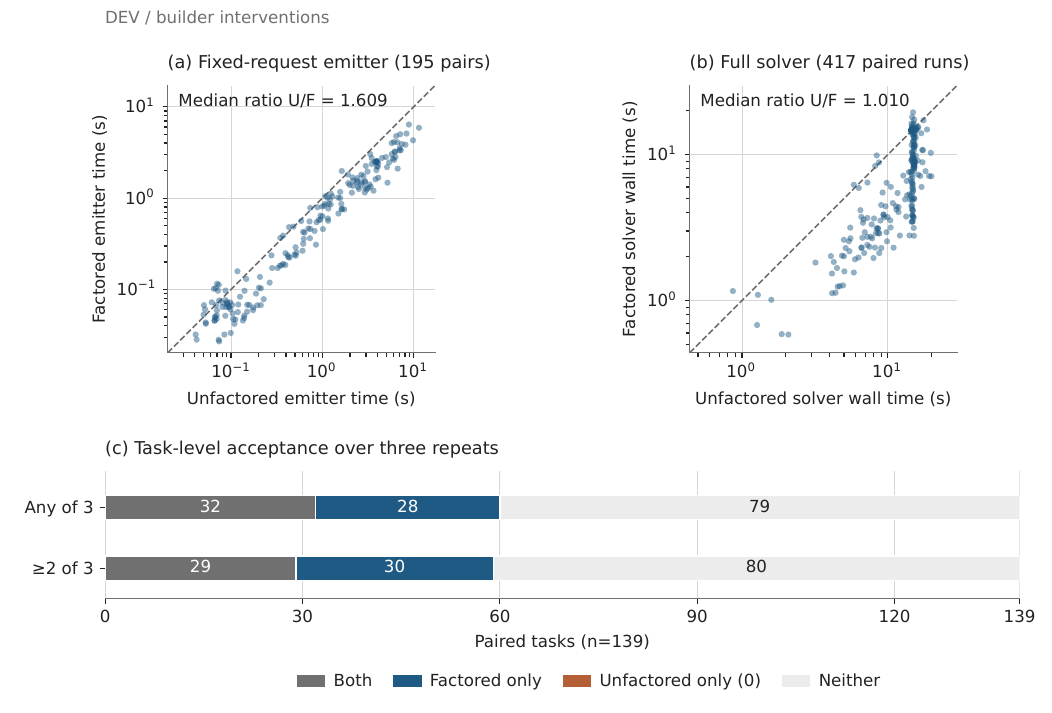}
\caption{Development evidence for factoring at two measurement scopes. (a) Paired emitter times for 65 success-biased fixed-request fixtures, each repeated three times. (b) Paired full-solver wall times for 139 tasks, each repeated three times. All pairs are plotted on equal logarithmic axes; the dashed line denotes equal time and U/F denotes unfactored divided by factored time. (c) Task-level acceptance in any repeat or in at least two of three repeats. The four categories partition the same 139 tasks.}
\label{fig:p1e2}
\end{figure*}

\textit{E4: terminal evidence improves the principal 15 s repair policy.} With no reserved repair slice and at most three repairs, reverified acceptance is 63/139 for full witnesses, 57 for position-filtered witnesses, 52 without repair, and 44 for blind translation. Full repair has 15 full-only and four no-repair-only outcomes, a net gain of 11 tasks or 7.91 percentage points. Full and position-filtered arms agree on 129/139 verdicts. Position filtering retains speed-label information, and the blind arm changes host proposal handling; these contrasts compare host policies rather than isolating every witness component.

The settings matrix in Appendix~\ref{app:p1details} records policy sensitivities. With the default reserve, blind translation exceeds no repair (59 versus 55); at 5 s, full repair reaches 21 versus 22 without repair. The one- and six-repair runs use a different worker count from the principal three-repair run, so cross-run changes combine cap and scheduling effects.

\subsection{Certificate Replay and Negative Controls}
Offline V08 replay accepts all 1,735 stored CUBI-DT certificates. Of these, 1,731 have outcome \texttt{CERTIFIED\_GOAL}; four have \texttt{VERIFIED\_AFTER\_BUDGET\_NOT\_SUCCESS} and remain online failures. A separate nominal body-sweep audit reports all 1,735 certificate curves clear of obstacles and boundaries. It uses linear translation and wrapped-heading interpolation between recorded poses, thereby checking nominal geometry rather than uncertainty-tube containment. The runtime import-path length remains 11 before and after every retained job, resolving the path-growth symptom recorded in an earlier series.

The three blocked task geometries reject in all 44 method/budget/tolerance settings, giving 132 negative job rows. These controls check rejection behavior on three distinct geometries. Same-kernel replay and nominal audits provide implementation consistency evidence; the conditional theorem additionally requires the enclosure and execution assumptions.

\subsection{LONG: Strict and Interface Outcomes}\label{sec:trackc}
Table~\ref{tab:trackc} reports LONG with both retained endpoints. The certificate-derived tracker achieves 90/140 strict successes on clean CUBI, 88/140 under disturbance, and 120/215 on BARN in each setting. The external stack records zero strict successes but 38/140, 41/140, 107/215, and 102/215 interface successes. The strict/interface split reflects the execution and observation interfaces described below.

\begin{table*}[!tp]
\caption{LONG route-tracking outcomes. Counts use every episode in each panel/setting. The tracker uses available long-budget certificates. Strict, interface, and recorded collision endpoints are shown together; action termination and native goal-checker differences are described in the text.}
\label{tab:trackc}\centering\small
\begin{tabular}{llrrrrrr}\toprule
&&\multicolumn{3}{c}{Certificate-derived route tracker}&\multicolumn{3}{c}{Lattice + MPPI + CM}\\
\cmidrule(lr){3-5}\cmidrule(lr){6-8}
Panel & Setting & Strict & Interface & Collisions & Strict & Interface & Collisions\\\midrule
CUBI28 & Clean & 90/140 & 90/140 & 0 & 0/140 & 38/140 & 11\\
CUBI28 & Disturbed & 88/140 & 88/140 & 2 & 0/140 & 41/140 & 9\\
BARN43 & Clean & 120/215 & 120/215 & 0 & 0/215 & 107/215 & 10\\
BARN43 & Disturbed & 120/215 & 120/215 & 0 & 0/215 & 102/215 & 6\\\bottomrule
\end{tabular}
\end{table*}

\begin{figure*}[!tp]
\centering\includegraphics[width=\textwidth]{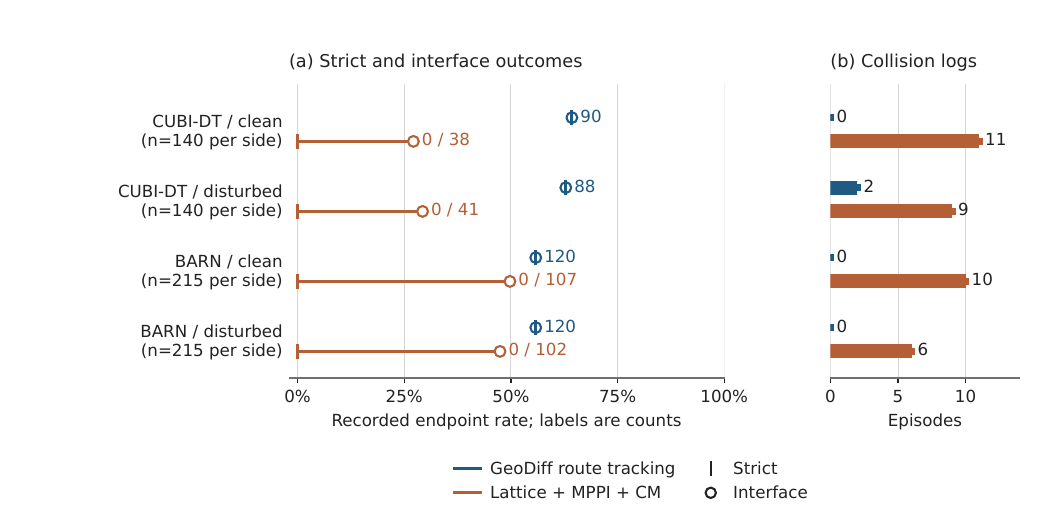}
\caption{Recorded LONG route-tracking outcomes. (a) A vertical mark denotes strict stopping and an open circle the interface endpoint; coincident marks indicate equal counts. Labels show strict/interface counts, or one count when equal. (b) Logged collisions. Each CUBI-DT condition contains 140 episodes per side and each BARN condition 215. Endpoint differences reflect the action-termination and goal-checker interfaces described in the text.}
\label{fig:long}
\end{figure*}

The retained external handler ends when the follow action finishes, before a common post-action stop-and-dwell window is guaranteed. In addition, the native goal-checker configuration uses positional tolerance $\max(\varepsilon_p,0.125\,\mathrm{m})$, whereas the strict test uses task fields. The actual generated configuration for each episode is unavailable. Consequently, the all-zero external strict column characterizes this observation procedure and cannot establish an intrinsic stopping disadvantage.

The tracker records two collisions on one disturbed CUBI task, in two seeds using a 60 s source certificate. Extracted-route tracking omits certified primitive execution information, and the retained simulator lacks independent swept-occupancy checking through every branch. These observations locate the gap between route following and the contract-faithful execution required by Theorem~\ref{thm:main}.

Restricting source certificates to 15 s in P0 reduces CUBI strict successes to 55/140 in both settings, while BARN remains 120/215 (Appendix~\ref{app:p0timing}). This identifies dependence on source-certificate availability. Scheduling and external outcomes also differ between the retained runs, so P0 is reported as a sensitivity rather than an isolated planning-budget effect.

\section{Discussion}\label{sec:discussion}
\subsection{What the Coverage Gain Explains}
GeoDiff connects the requested stopping condition to the allocation of geometric computation. Original-goal inclusion supplies an explicit acceptance target. Terminal witnesses turn a narrowly failed stopping condition into supported repair candidates. Factoring reduces repeated work across overlapping proof obligations. In the primary CUBI-DT condition, their combination yields 17 more certificates than Stop-set within the same recorded budget.

The paired results and deadline reversals explain the nature of this gain. GeoDiff increases aggregate coverage while exchanging some successes with its references; repair competes with continued search for the remaining budget. E2 independently identifies savings at the emitter and higher repeated-attempt coverage at the solver. These results favor certification-aware organization of search without requiring a uniform end-to-end speedup or task-wise dominance.

\subsection{Scope and Transfer}
The formulation accommodates finite unions of convex body parts, shared-error representations, and alternative proposal sources. Transfer requires conservative primitive enclosures and a controller satisfying the corresponding physical contract. The four BARN-derived conditions broaden the geometric evaluation across body models and tolerances, while retaining a certificate endpoint. CUBI-DT remains an intentionally focused, scale-normalized door-and-turn distribution; staged inspection and prior BARN exposure support descriptive interpretation.

The execution boundary is concrete. LONG follows extracted routes, uses available source certificates from unequal planning budgets, and retains different terminal observation behavior across systems. A faithful certificate-program executor or a demonstrated refinement relation is needed to apply the model-level guarantee to that controller. The two recorded tracker collisions and strict/interface discrepancies are therefore presented alongside successes. Complete frozen task contracts, executor configurations, and uniform process-level timing would support stronger reproducibility and portable runtime claims; they are not supplied by the aggregate evidence artifact.

\section{Conclusion}\label{sec:conclusion}
GeoDiff makes a certifiable stopping program the target of finite-budget search. It preserves the original goal, uses an exact fixed-enclosure characterization to guide terminal repair, and reuses repeated proof expressions under explicit equivalence conditions. The primary evaluation increases certification from 44/139 to 61/139 tasks within 15 s, with complete results exposing how deadline, tolerance, and repair allocation affect coverage. Public-map tasks and development measurements extend this account of the method, while LONG identifies the controller-contract link required for faithful execution. The resulting formulation offers a concrete way to organize planning around both geometric feasibility and the computational cost of establishing it.

\appendix
\section{Complete Budget and Tolerance Matrix}\label{app:counts}
Table~\ref{tab:all} gives the full positive-task matrix. Negative controls are reported separately to avoid diluting success-rate denominators with tasks designed to be blocked. The data and recount program retain the original task and cluster identifiers, all failures, and the four late certificates. Reproducing these aggregate numbers requires only the recorded rows; reproducing the planner requires the frozen task geometry, kernel, candidate generator, and execution configuration as well.
\begin{table*}[!tp]
\caption{Complete certified-success counts. Each 0.9 m column has 139 positive tasks; each 0.8 m column has 118. Budgets are seconds and goal tolerances are meters.}
\label{tab:all}\centering\small
\begin{tabular}{lrrrrrrrr}\toprule
&\multicolumn{4}{c}{0.9 m normalization ($n=139$)}&\multicolumn{4}{c}{0.8 m normalization ($n=118$)}\\
\cmidrule(lr){2-5}\cmidrule(lr){6-9}
&\multicolumn{2}{c}{0.05 goal}&\multicolumn{2}{c}{0.15 goal}&\multicolumn{2}{c}{0.05 goal}&\multicolumn{2}{c}{0.15 goal}\\
Method&5 s&15 s&5 s&15 s&5 s&15 s&5 s&15 s\\\midrule
\allrows
\bottomrule\end{tabular}
\end{table*}

\section{Development Measurement Details}\label{app:p1details}
All E1/E2/E4 measurements in this appendix are development-only. The observed native-found/convertible/strict chain is distinct from the inferred geometric and brake indicators. In the external rows, inferred geometric/brake counts are 37/43 for Lattice B+, 14/19 for Hybrid B+, and 21/22 for DDR B+. These non-nested values cannot be interpreted as survivor counts of consecutive proof stages.

The E2 emitter ratios compare the same immutable request. The 65 fixtures are explicitly success-biased and have three repetitions; all 195 recorded certificate hashes, semantic hashes, and verdicts agree. In full-task E2, the legacy \texttt{compile\_s} field instead times the entire solve, and its semantic hash contains the builder identifier. Consequently, a failed aggregate hash-equivalence flag does not establish that memoization changes mathematical semantics. The 60-versus-32 any-of-three result remains a descriptive task aggregate, not a single-attempt deployment success rate.

Table~\ref{tab:e4settings} gives the E4 settings and acceptances. The endpoint is reverified acceptance under a solver-budget setting. Adding a filter for logged solve-plus-reverification time at most 15 s would give 46/38/46/55 in the principal none/blind/position-filtered/full row, rather than 52/44/57/63. This timer-scope sensitivity does not replace the frozen TEST predicate. Repair metadata also contain placeholders rather than complete per-candidate shift/label/interval observations; the paired full-only outcomes do not independently identify a causal terminal shift.

\begin{table*}[!tp]
\caption{P1 E4 development settings and reverified acceptances (139 tasks per arm). Cap is the repair-policy bound, not an equal realized candidate count. Missing short-budget policy metadata remain unspecified.}
\label{tab:e4settings}\centering\small
\begin{tabular}{lrrrrrrrr}\toprule
Setting & Solver (s) & Reserve & Cap & Workers & None & Blind & Pos.-filtered & Full\\\midrule
Default &15&0.2&3&20&55&59&69&70\\
Fair principal &15&0&3&28&52&44&57&63\\
Fair cap 1 &15&0&1&14&39&37&42&44\\
Fair cap 6 &15&0&6&14&40&31&42&43\\
Short budget &5&--&--&14&22&19&22&21\\\bottomrule
\end{tabular}
\end{table*}

Table~\ref{tab:e4settings} retains every setting, including different worker counts and changing no-repair outcomes. Position filtering retains speed labels. The within-setting contrasts support the policy comparisons in Section~\ref{sec:p1}; cross-row changes combine multiple configuration factors.

\section{15 s Source-Certificate Sensitivity and Timing}\label{app:p0timing}
Table~\ref{tab:p0inpaper} reports every P0 strict/interface/collision endpoint using 15 s source certificates. Its episodes remain separate from LONG to preserve the certificate-selection sensitivity.

\begin{table*}[!tp]
\caption{P0 sensitivity: 15 s source certificates only. Strict and interface outcomes are retained together; collision counts are recorded events. The same execution and observation limitations discussed for LONG apply.}
\label{tab:p0inpaper}\centering\small
\begin{tabular}{llrrrrrr}\toprule
&&\multicolumn{3}{c}{Certificate-derived route tracker}&\multicolumn{3}{c}{Lattice + MPPI + CM}\\
\cmidrule(lr){3-5}\cmidrule(lr){6-8}
Panel&Setting&Strict&Interface&Collisions&Strict&Interface&Collisions\\\midrule
CUBI28&Clean&55/140&55/140&0&0/140&42/140&5\\
CUBI28&Disturbed&55/140&55/140&0&0/140&38/140&10\\
BARN43&Clean&120/215&120/215&0&0/215&101/215&9\\
BARN43&Disturbed&120/215&120/215&0&0/215&102/215&9\\\bottomrule
\end{tabular}
\end{table*}

Figure~\ref{fig:testtime} includes every positive primary task in the denominator. The curves summarize accepted, on-time certificates from a 15 s run and do not assume that a separate 5 s run follows the same search prefix. The recorded process-boundary asymmetry precludes a hardware-independent runtime ranking.

\section{Reproducibility and Artifact Scope}\label{app:repro}
The analysis artifact contains retained job and episode records, recount scripts, figure inputs, and a claim-to-evidence ledger. The supplied freeze metadata binds the primary campaign. Recorded rows retain body dimensions, map residual, normalization, goal tolerances, and budgets. The artifact reproduces aggregate statistics and plots. Complete end-to-end reconstruction additionally requires the original task-level plant/shared-error parameters, task geometries, core planner/verifier, raw certificates, full trajectories, and per-episode executor configurations; these are not all present in the retained package. Host defaults are not substituted for missing frozen fields.

All online counts preserve recorded outcome semantics: 1,731 CUBI-DT outcomes are \texttt{CERTIFIED\_GOAL}; four late certificates remain failures. P1 retains its development role despite exposed-task reuse, and P0 remains a selection sensitivity. The public-map replay summary separately accounts for 829 accepted available certificates and 12 records without a certificate. Mathematical results specify conditional invariants; replay and nominal sweeps supply the recorded implementation checks.

\section*{Acknowledgment}
OpenAI ChatGPT assisted with manuscript restructuring, language editing, mathematical exposition, literature organization, and plotting and caption code. Generative image assistance supplied conceptual visual motifs for Fig.~\ref{fig:method_architecture}; the scientific connections and labels were composed as vector overlays. Figs.~\ref{fig:shared_uncertainty} and~\ref{alg:repair} were typeset as vector diagrams using the stated analytical example and documented policy, respectively. Quantitative results and plots are based on retained records, not generated images. The authors retain responsibility for the claims, proofs, references, figures, and interpretation.

\bibliographystyle{elsarticle-num}
\bibliography{cas-refs}
\end{document}